\documentclass[11pt,letterpaper]{llncs}

\usepackage{amsmath,amssymb}
\usepackage{graphicx}
\usepackage{marvosym}
\usepackage{url}
\usepackage{fancyhdr}
\usepackage{geometry}
\usepackage[utf8]{inputenc}
\usepackage[T1]{fontenc}

\usepackage{microtype}
\usepackage{times}

\usepackage{amsfonts}
\usepackage{bbm}
\usepackage{nicefrac}
\usepackage{array}
\usepackage{multirow}
\usepackage{enumitem}
\usepackage{booktabs}

\usepackage[ruled,vlined,linesnumbered]{algorithm2e}
\usepackage{algorithmic}

\usepackage[accsupp]{axessibility}
\usepackage{float}
\usepackage{adjustbox}
\usepackage{caption}

\usepackage[dvipsnames]{xcolor}
\usepackage{hyperref}

\usepackage{epsfig}

\newcommand{\ours}{\mathrm{ExplainSeg}\xspace}

\newcommand{\keywords}[1]{\par\addvspace\baselineskip
\noindent\keywordname\enspace\ignorespaces#1}

\fancypagestyle{firstpage}{\fancyhf{}
\fancyhead[C]{\small{International Conference on Digital Image Processing and Vision (ICDIPV) }}
\rfoot{\thepage}
}

\begin{document}

\def\x{{\mathbf x}}
\def\L{{\cal L}}
\title{\LARGE{No Masks Needed: Explainable AI for Deriving Segmentation from Classification}}


%
%
\author{\large{Mosong Ma\textsuperscript{1} \and Tania Stathaki\textsuperscript{1} \and Michalis Lazarou\textsuperscript{2}}}
\institute{\textsuperscript{1}\large{Dept. Electrical and Electronic Engineering, Imperial College London,\\ London, UK}
\\
\textsuperscript{2}\large{Centre for Vision, Speech and Signal Processing, University of Surrey,\\ Guildford, UK}}

%
%


\newcommand{\head}[1]{{\medskip\noindent\textbf{#1}\hspace{10pt}}}
\newcommand{\alert}[1]{{\color{red}{#1}}}
\newcommand{\sm}{\scriptsize}
\newcommand{\eq}[1]{(\ref{eq:#1})}

\newcommand{\Th}[1]{\textsc{#1}}
\newcommand{\mr}[2]{\multirow{#1}{*}{#2}}
\newcommand{\mc}[2]{\multicolumn{#1}{c}{#2}}
\newcommand{\tb}[1]{\textbf{#1}}
\newcommand{\ch}{\checkmark}

\newcommand{\red}[1]{{\color{red}{#1}}}
\newcommand{\blue}[1]{{\color{blue}{#1}}}
\newcommand{\green}[1]{{\color{green}{#1}}}
\newcommand{\gray}[1]{{\color{gray}{#1}}}

\newcommand{\citeme}[1]{\red{[XX]}}
\newcommand{\refme}[1]{\red{(XX)}}


\newcommand{\tran}{^\top}
\newcommand{\mtran}{^{-\top}}
\newcommand{\zcol}{\mathbf{0}}
\newcommand{\zrow}{\zcol\tran}

\newcommand{\ind}{\mathbbm{1}}
\newcommand{\expect}{\mathbb{E}}
\newcommand{\nat}{\mathbb{N}}
\newcommand{\zahl}{\mathbb{Z}}
\newcommand{\real}{\mathbb{R}}
\newcommand{\proj}{\mathbb{P}}
\newcommand{\prob}{\mathbf{Pr}}
\newcommand{\normal}{\mathcal{N}}

\newcommand{\mif}{\textrm{if}\ }
\newcommand{\other}{\textrm{otherwise}}
\newcommand{\minimize}{\textrm{minimize}\ }
\newcommand{\maximize}{\textrm{maximize}\ }
\newcommand{\st}{\textrm{subject\ to}\ }

\newcommand{\id}{\operatorname{id}}
\newcommand{\const}{\operatorname{const}}
\newcommand{\sgn}{\operatorname{sgn}}
\newcommand{\var}{\operatorname{Var}}
\newcommand{\mean}{\operatorname{mean}}
\newcommand{\trace}{\operatorname{tr}}
\newcommand{\diag}{\operatorname{diag}}
\newcommand{\vect}{\operatorname{vec}}
\newcommand{\cov}{\operatorname{cov}}
\newcommand{\sign}{\operatorname{sign}}
\newcommand{\prj}{\operatorname{proj}}

\newcommand{\softmax}{\operatorname{softmax}}
\newcommand{\clip}{\operatorname{clip}}

\newcommand{\defn}{\mathrel{:=}}
\newcommand{\peq}{\mathrel{+\!=}}
\newcommand{\meq}{\mathrel{-\!=}}

\newcommand{\floor}[1]{\left\lfloor{#1}\right\rfloor}
\newcommand{\ceil}[1]{\left\lceil{#1}\right\rceil}
\newcommand{\inner}[1]{\left\langle{#1}\right\rangle}
\newcommand{\norm}[1]{\left\|{#1}\right\|}
\newcommand{\abs}[1]{\left|{#1}\right|}
\newcommand{\frob}[1]{\norm{#1}_F}
\newcommand{\card}[1]{\left|{#1}\right|\xspace}
\newcommand{\diff}{\mathrm{d}}
\newcommand{\der}[3][]{\frac{d^{#1}#2}{d#3^{#1}}}
\newcommand{\pder}[3][]{\frac{\partial^{#1}{#2}}{\partial{#3^{#1}}}}
\newcommand{\ipder}[3][]{\partial^{#1}{#2}/\partial{#3^{#1}}}
\newcommand{\dder}[3]{\frac{\partial^2{#1}}{\partial{#2}\partial{#3}}}

\newcommand{\wb}[1]{\overline{#1}}
\newcommand{\wt}[1]{\widetilde{#1}}

\def\xssp{\hspace{1pt}}
\def\ssp{\hspace{3pt}}
\def\msp{\hspace{5pt}}
\def\lsp{\hspace{12pt}}

\newcommand{\cA}{\mathcal{A}}
\newcommand{\cB}{\mathcal{B}}
\newcommand{\cC}{\mathcal{C}}
\newcommand{\cD}{\mathcal{D}}
\newcommand{\cE}{\mathcal{E}}
\newcommand{\cF}{\mathcal{F}}
\newcommand{\cG}{\mathcal{G}}
\newcommand{\cH}{\mathcal{H}}
\newcommand{\cI}{\mathcal{I}}
\newcommand{\cJ}{\mathcal{J}}
\newcommand{\cK}{\mathcal{K}}
\newcommand{\cL}{\mathcal{L}}
\newcommand{\cM}{\mathcal{M}}
\newcommand{\cN}{\mathcal{N}}
\newcommand{\cO}{\mathcal{O}}
\newcommand{\cP}{\mathcal{P}}
\newcommand{\cQ}{\mathcal{Q}}
\newcommand{\cR}{\mathcal{R}}
\newcommand{\cS}{\mathcal{S}}
\newcommand{\cT}{\mathcal{T}}
\newcommand{\cU}{\mathcal{U}}
\newcommand{\cV}{\mathcal{V}}
\newcommand{\cW}{\mathcal{W}}
\newcommand{\cX}{\mathcal{X}}
\newcommand{\cY}{\mathcal{Y}}
\newcommand{\cZ}{\mathcal{Z}}

\newcommand{\vA}{\mathbf{A}}
\newcommand{\vB}{\mathbf{B}}
\newcommand{\vC}{\mathbf{C}}
\newcommand{\vD}{\mathbf{D}}
\newcommand{\vE}{\mathbf{E}}
\newcommand{\vF}{\mathbf{F}}
\newcommand{\vG}{\mathbf{G}}
\newcommand{\vH}{\mathbf{H}}
\newcommand{\vI}{\mathbf{I}}
\newcommand{\vJ}{\mathbf{J}}
\newcommand{\vK}{\mathbf{K}}
\newcommand{\vL}{\mathbf{L}}
\newcommand{\vM}{\mathbf{M}}
\newcommand{\vN}{\mathbf{N}}
\newcommand{\vO}{\mathbf{O}}
\newcommand{\vP}{\mathbf{P}}
\newcommand{\vQ}{\mathbf{Q}}
\newcommand{\vR}{\mathbf{R}}
\newcommand{\vS}{\mathbf{S}}
\newcommand{\vT}{\mathbf{T}}
\newcommand{\vU}{\mathbf{U}}
\newcommand{\vV}{\mathbf{V}}
\newcommand{\vW}{\mathbf{W}}
\newcommand{\vX}{\mathbf{X}}
\newcommand{\vY}{\mathbf{Y}}
\newcommand{\vZ}{\mathbf{Z}}

\newcommand{\va}{\mathbf{a}}
\newcommand{\vb}{\mathbf{b}}
\newcommand{\vc}{\mathbf{c}}
\newcommand{\vd}{\mathbf{d}}
\newcommand{\ve}{\mathbf{e}}
\newcommand{\vf}{\mathbf{f}}
\newcommand{\vg}{\mathbf{g}}
\newcommand{\vh}{\mathbf{h}}
\newcommand{\vi}{\mathbf{i}}
\newcommand{\vj}{\mathbf{j}}
\newcommand{\vk}{\mathbf{k}}
\newcommand{\vl}{\mathbf{l}}
\newcommand{\vm}{\mathbf{m}}
\newcommand{\vn}{\mathbf{n}}
\newcommand{\vo}{\mathbf{o}}
\newcommand{\vp}{\mathbf{p}}
\newcommand{\vq}{\mathbf{q}}
\newcommand{\vr}{\mathbf{r}}
\newcommand{\Vs}{\mathbf{s}}
\newcommand{\vt}{\mathbf{t}}
\newcommand{\vu}{\mathbf{u}}
\newcommand{\vv}{\mathbf{v}}
\newcommand{\vw}{\mathbf{w}}
\newcommand{\vx}{\mathbf{x}}
\newcommand{\vy}{\mathbf{y}}
\newcommand{\vz}{\mathbf{z}}

\newcommand{\vone}{\mathbf{1}}
\newcommand{\vzero}{\mathbf{0}}

\newcommand{\valpha}{{\boldsymbol{\alpha}}}
\newcommand{\vbeta}{{\boldsymbol{\beta}}}
\newcommand{\vgamma}{{\boldsymbol{\gamma}}}
\newcommand{\vdelta}{{\boldsymbol{\delta}}}
\newcommand{\vepsilon}{{\boldsymbol{\epsilon}}}
\newcommand{\vzeta}{{\boldsymbol{\zeta}}}
\newcommand{\veta}{{\boldsymbol{\eta}}}
\newcommand{\vtheta}{{\boldsymbol{\theta}}}
\newcommand{\viota}{{\boldsymbol{\iota}}}
\newcommand{\vkappa}{{\boldsymbol{\kappa}}}
\newcommand{\vlambda}{{\boldsymbol{\lambda}}}
\newcommand{\vmu}{{\boldsymbol{\mu}}}
\newcommand{\vnu}{{\boldsymbol{\nu}}}
\newcommand{\vxi}{{\boldsymbol{\xi}}}
\newcommand{\vomikron}{{\boldsymbol{\omikron}}}
\newcommand{\vpi}{{\boldsymbol{\pi}}}
\newcommand{\vrho}{{\boldsymbol{\rho}}}
\newcommand{\vsigma}{{\boldsymbol{\sigma}}}
\newcommand{\vtau}{{\boldsymbol{\tau}}}
\newcommand{\vupsilon}{{\boldsymbol{\upsilon}}}
\newcommand{\vphi}{{\boldsymbol{\phi}}}
\newcommand{\vchi}{{\boldsymbol{\chi}}}
\newcommand{\vpsi}{{\boldsymbol{\psi}}}
\newcommand{\vomega}{{\boldsymbol{\omega}}}

\newcommand{\rLambda}{\mathrm{\Lambda}}
\newcommand{\rSigma}{\mathrm{\Sigma}}

\newcommand{\p}{\mathrm{\mathrm{p}}}
\newcommand{\z}{\mathrm{\mathrm{z}}}
\newcommand{\y}{\mathrm{\mathrm{y}}}

\newcommand{\vLambda}{\bm{\rLambda}}
\newcommand{\vSigma}{\bm{\rSigma}}

\makeatletter
\newcommand*\bdot{\mathpalette\bdot@{.7}}
\newcommand*\bdot@[2]{\mathbin{\vcenter{\hbox{\scalebox{#2}{$\m@th#1\bullet$}}}}}
\makeatother

\makeatletter
\DeclareRobustCommand\onedot{\futurelet\@let@token\@onedot}
\def\@onedot{\ifx\@let@token.\else.\null\fi\xspace}

\def\eg{\emph{e.g}\onedot} \def\Eg{\emph{E.g}\onedot}
\def\ie{\emph{i.e}\onedot} \def\Ie{\emph{I.e}\onedot}
\def\cf{\emph{cf}\onedot} \def\Cf{\emph{Cf}\onedot}
\def\etc{\emph{etc}\onedot} \def\vs{\emph{vs}\onedot}
\def\wrt{w.r.t\onedot} \def\dof{d.o.f\onedot} \def\aka{a.k.a\onedot}
\def\etal{\emph{et al}\onedot}
\makeatother

\maketitle

\begin{abstract}
\emph{Medical image segmentation is vital for modern healthcare and is a key element of computer-aided diagnosis. While recent advancements in computer vision have explored unsupervised segmentation using pre-trained models, these methods have not been translated well to the medical imaging domain. In this work, we introduce a novel approach that fine-tunes pre-trained models specifically for medical images, achieving accurate segmentation with extensive processing. Our method integrates Explainable AI to generate relevance scores, enhancing the segmentation process. Unlike traditional methods that excel in standard benchmarks but falter in medical applications, our approach achieves improved results on datasets like CBIS-DDSM, NuInsSeg and Kvasir-SEG.}
\keywords{\emph{Medical Image Segmentation, Explainable AI, Transfer Learning.}}
\end{abstract}

\section{Introduction}
\label{sec:intro}

Medical image segmentation has become a focal point in research due to its potential to revolutionize healthcare by enabling faster and more accurate diagnoses \cite{medical_deep_survey}. This task, however, is challenging, particularly because deep learning methods rely heavily on large, well-annotated datasets. In the medical domain, such datasets are difficult to obtain because of privacy concerns and the need for expert-level annotations. These annotations require precise pixel-level labelling, which is time-consuming and costly, often demanding physicians to divert time from patient care.

To overcome these challenges, recent approaches have explored using weakly supervised learning \cite{weaklysupervised} with weak annotations or applying transfer learning \cite{transferlearning} from pre-trained models. Our work focuses on the latter, building on the idea of fine-tuning pre-trained classification models specifically for medical tasks. While prior studies \cite{tokencut,cutler,oriane_unsupervised} have used pre-trained models for segmentation without fine-tuning, achieving notable results in standard image datasets, these methods struggle or fail in medical image segmentation.

\begin{figure}[!ht]
    \centering
    \includegraphics[width=0.8\columnwidth]{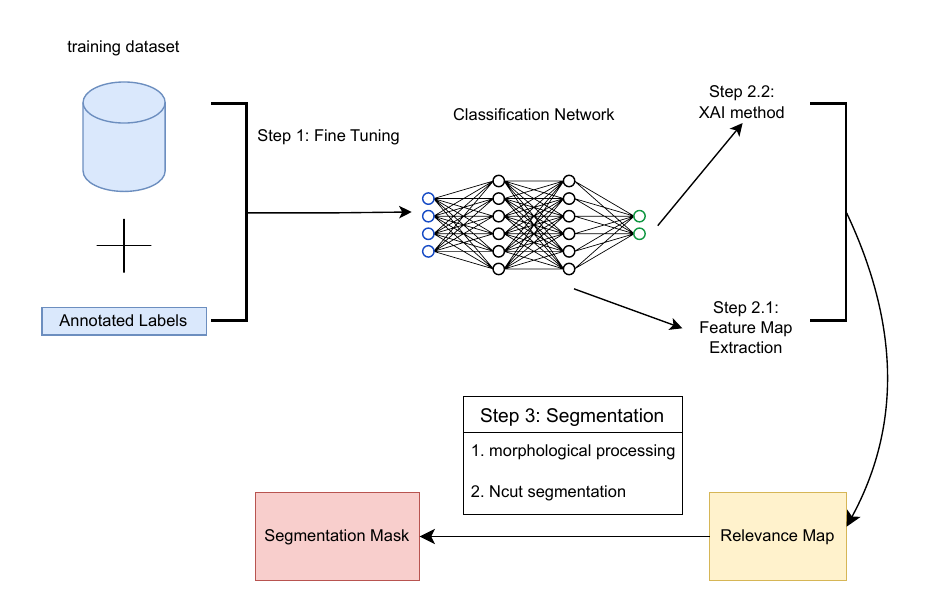}
    \caption{General Pipeline for ExplainSeg}
    \label{fig: general pipeline}
\end{figure}

Our approach addresses these limitations by fine-tuning a linear classifier on the medical dataset and utilizing Explainable Artificial Intelligence (XAI) \cite{XAI} to interpret the model's decision-making process. This interpretation is converted into segmentation masks, providing a more human-readable and clinically useful output. By doing so, we demonstrate that fine-tuning, combined with XAI, can enhance the performance of pre-trained models in the medical imaging domain, leading to more effective and interpretable segmentation results.

Our contributions are summarized below:
\begin{itemize}
    \item We propose a novel method for medical image segmentation ($\ours$).
    \item We show the efficacy of XAI in medical segmentation by simply fine-tuning a linear probing classifier on top of a pre-trained model
    \item $\ours$ achieves state-of-the-art performance and significantly outperforms other segmentation methods using pre-trained medical image segmentation models.
\end{itemize}

\section{Related work}

\subsection{Medical image Segmentation}
Medical image segmentation \cite{MIS} has been widely studied, with various methods developed to address its challenges. Early approaches relied on manual or semi-automated techniques, which require significant human input, making them time-consuming and prone to variability between observers, limiting their use in large-scale studies. Segmentation methods \cite{ms} can be broadly categorized into traditional techniques, such as edge-based, region-based, and thresholding \cite{handbook}, along with machine learning methods, which include supervised, semi-supervised, and unsupervised approaches.

\subsection{Transfer Learning Segmentation}
\label{sec:fsl}
Transfer learning \cite{transferlearning} involves leveraging knowledge from a pre-trained model, typically trained on a large dataset, and applying it to a different but related task. This approach has become popular in image segmentation tasks, particularly when labelled data is scarce or models must generalize across different domains.

Several methods have successfully applied transfer learning to segmentation tasks. TokenCut \cite{tokencut} is a notable example, where a Vision Transformer (ViT) \cite{vit} is pre-trained through the self-supervised learning method of DINO \cite{DINO} and used to segment objects by treating image patches as tokens and optimizing a normalized cut criterion. Cut-and-LEaRn (CutLER) \cite{cutler} and FOUND \cite{oriane_unsupervised} also employ pre-trained networks to analyze background regions and generate coarse object proposals.

\subsection{Explainable Artificial Intelligence (XAI)}
\label{sec:transductive_fsl}

Machine learning has significantly advanced medical image segmentation, with deep learning models like U-Net becoming the standard for their ability to learn complex features. However, these models are often viewed as "black boxes" due to their opaque decision-making processes.

Explainable AI (XAI) is increasingly vital in addressing this issue. Techniques like integrated gradients \cite{IG} and Grad-CAM \cite{gradcam} provide visual explanations, helping users understand which parts of an image influenced the model's predictions. While these tools enhance the balance between performance and interpretability, challenges remain in ensuring that these explanations are accurate and relevant.

\subsection{XAI to Segmentation}

Our approach leverages deep learning and Explainable AI (XAI) techniques to enhance interpretability and reduce reliance on pixel-level annotations in medical image segmentation. Using transfer learning, we adapt a pre-trained model to our dataset and employ XAI to generate segmentation masks, ensuring both accuracy and transparency.

Previous works, such as \cite{e2s} and  \cite{c2s}, leverage XAI to create segmentation masks without pixel-level labels in generic domains like infrastructure imagery. In contrast, our work focuses on the medical imaging domain, addressing its higher precision and robustness requirements. By tailoring XAI methods to weakly annotated medical data, we demonstrate their potential to deliver accurate and meaningful segmentation results, highlighting the novelty of our approach.

\section{Method}
\label{sec:method}

\subsection{Problem formulation}
\label{sec:problem_formulation}

At the fine-tuning stage, we assume access to a labelled dataset, $D_{\text{train}} = \{\vx_i, y_i\}_{i=1}^N$,  of $N$ image-label pairs, where every image $\vx_{i}$ has a label $y_i \in Y$, that belongs to the label set $Y$. We fine-tune a classification network $F$ on $D_{\text{train}}$.

This fine-tuned network $F$ is then trained with an extra classification head $G$ to generate and interpret classification results with the help of XAI. The generated relevance map is refined into segmentation masks with extensive post-processing including morphology and other segmentation methods.

\subsection{Optional Fine Tuning \& Training}
\label{sec:features}

When working with a new dataset, fine-tuning \cite{finetune} is necessary if a pre-trained classification network struggles to classify accurately. To minimize computational demands, gradients are typically frozen for most of the hidden layers. A task-specific classification layer can also be added on top of the deep learning model, and its parameters are trainable allowing the model to adapt to the specific medical image dataset.

\subsection{Integrated Gradients (IG)}
Integrated gradients \cite{IG} trace how the network determines an image's class by assigning relevance scores to each pixel that indicate its contribution to the classification result, and is chosen due to its model agnostic feature making it compatible to most models.

The integrated gradient along the $i$-th dimension for an input $\mathbf{x}$ and baseline $\mathbf{x}'$ is defined as follows:

\begin{equation}
\text{IG}_i(\mathbf{x}) = (\mathbf{x}_i - \mathbf{x}'_i) \times \int_{\alpha=0}^{1} \frac{\partial F(\mathbf{x}' + \alpha \times (\mathbf{x} - \mathbf{x}'))}{\partial \mathbf{x}_i} \, d\alpha
\label{eq:IG}
\end{equation}

where \( \alpha \) is a scaling factor that interpolate between the baseline \( \mathbf{x}' \) and the input \( \mathbf{x} \), $\frac{\partial F(\mathbf{x})}{\partial x_i}$ is the gradient of $F(\mathbf{x})$ along the $i$-th dimension.

The Noise Tunnel function in Captum \cite{captum} is an extension of existing attribution methods used for explainable AI (XAI) by adding stochastic noise to input samples to enhance the robustness and reliability of the explanations. It applies Gaussian noise to the input data in multiple iterations, averaging the resulting attribution maps. This process helps reduce the model sensitivity to small perturbations and highlights consistent patterns in the attributions. As a result, Noise Tunnel produces more stable and interpretable explanations.

We did not employ other XAI techniques such as Grad-CAM, as preliminary experiments have consistently shown that the coarse heatmaps produced by Grad-CAM offer significantly inferior spatial localization compared to IG, particularly for fine-grained structures in medical images. In our experiments, Grad-CAM failed to highlight diagnostically relevant regions with sufficient resolution or fidelity. Given these limitations and the superior performance of IG in our setting, further exploration of Grad-CAM and related methods was deemed unnecessary.

\subsection{Feature Map Extraction}
Alongside integrated gradients for relevance scores, we also explored extracting and visualizing feature maps from the ViT network \cite{featuremap}.

Feature map extraction is a key process in deep learning. It captures the essential patterns from an input image. As the image passes through the layers of a network, each layer applies filters that detect specific features and identify patterns relevant to tasks like object recognition or segmentation.

The result of this filtering process is the feature map, which highlights the areas of the image where the features were detected. Feature maps play a crucial role in helping the network understand the content of an image and perform tasks like classification or localization.

\subsection{Image Segmentation}
\label{sec:lp}

The relevance maps generated by XAI methods are insufficient for accurate segmentation mask generation. Therefore, additional post-processing is necessary to refine them into precise segmentation masks.

\head{Otsu Thresholding} The simplest approach involves applying thresholding to the IG results. Otsu thresholding \cite{otsu} is a method used to automatically find the optimal threshold for separating an image into two classes: foreground and background. It works by analyzing the histogram of pixel intensities and selecting the threshold that minimizes the intra-class variance, which measures the spread of pixel values within each class. This approach ensures that the selected threshold effectively distinguishes between the two groups of pixels.

In our application, we use the relevance map generated from the Integrated Gradients method as input for Otsu thresholding. The relevance map highlights the importance of each pixel in the input image for the prediction of the model. By applying Otsu thresholding to this map, we can filter out less relevant pixels and focus on the most significant areas that contributed to the model's decision. This helps reduce noise and emphasize the most pertinent regions in the relevance map.

\head{Morphological Processing} We applied a series of image processing and morphological operations to refine the pixel relevance maps generated by the Integrated Gradients method. First, the relevance maps were converted to grayscale and enhanced through histogram equalization \cite{adapthist}. Thresholding was then used to create binary masks that highlight significant regions. We labelled connected components, removed border artefacts, and used dilation to connect fragmented areas. Erosion was applied to clean the boundaries, and small holes were filled to produce smooth, continuous masks.

\head{Normalized Cut} The Normalized Cut method \cite{ncut} models an image as a weighted graph \( G = (V, E) \), where each pixel corresponds to a node \( v \in V \) and edges \( e \in E \) connect pairs of nodes. The weight \( w_{ij} \) of an edge between nodes \( i \) and \( j \) represents the similarity between the corresponding pixels.

The goal is to partition the graph into disjoint sets \( A \) and \( B \) by minimizing the Ncut value and therefore separating foreground and background, which can be defined as:
\begin{equation}
\text{NCut}(A, B) = \frac{\text{Cut}(A, B)}{\text{Assoc}(A, V)} + \frac{\text{Cut}(A, B)}{\text{Assoc}(B, V)}
\label{eq:Ncut}
\end{equation}

where:
\begin{equation}
\text{Cut}(A, B) = \sum_{i \in A, j \in B} w_{ij}
\end{equation}
\begin{equation}
\text{Assoc}(A, V) = \sum_{i \in A} \sum_{j \in V} w_{ij}, \text{Assoc}(B, V) = \sum_{i \in B} \sum_{j \in V} w_{ij}
\end{equation}

\head{Fusion of Feature and Relevance Map} Extracting the feature map from the Vision Transformer network enhances the XAI segmentation by revealing how the network perceives medical images. By integrating pixel-level importance with the feature space representation, we can reduce the influence of outlier pixels in the integrated gradient map. We combine the relevance map and feature map by multiplying them, effectively using the feature map as a weighted map to denoise the relevance map.

\subsection{DenseCRF Edge Refinement}
\label{sec:adapting_anchors}

Dense Conditional Random Field (DenseCRF) \cite{densecrf} is a powerful post-processing technique used to refine coarse segmentation masks, particularly by leveraging the information from the original image. They work by incorporating both the initial segmentation predictions and the pixel-level details of the original image to produce more accurate and coherent results.

DenseCRF models combine unary potentials, which come from the initial segmentation output and indicate the likelihood of each pixel belonging to a specific class, with pairwise potentials that consider the relationships between pixels based on their spatial location and appearance (colour, texture). The key advantage of DenseCRF is its ability to enforce pixels with similar appearances and close proximity in the original image are more likely to share the same label in the refined segmentation.

The energy function in DenseCRF is defined as:
\begin{equation}
E(\mathbf{x}) = \sum_{i} \phi_{u}(\mathbf{x}_{i}) + \sum_{i < j} \phi_{p}(\mathbf{x}_{i}, \mathbf{x}_{j})
\label{eq:densecrf}
\end{equation}

where \(\phi_{u}(\mathbf{x}_{i})\)  is the unary potential that captures the likelihood of pixel \( \mathbf{x}_{i} \) belonging to a particular class, and \(\phi_{p}(\mathbf{x}_{i}, \mathbf{x}_{j})\) is the pairwise potential that models the interactions between all neighbouring pixel pairs \( \mathbf{x}_{i} \) and \( \mathbf{x}_{j} \) without repetition.

Algorithm \ref{alg:adaptive_lp} summarizes our method.
\begin{algorithm}
\footnotesize
\DontPrintSemicolon
\SetFuncSty{textsc}
\SetDataSty{emph}
\newcommand{\commentsty}[1]{{\color{Blue}#1}}
\SetCommentSty{commentsty}
\SetKwComment{Comment}{$\triangleright$ }{}
\SetKwInOut{Input}{input}
\SetKwInOut{Output}{output}
\SetKwFunction{Graph}{graph}
\SetKwFunction{Label}{label}
\SetKwFunction{Power}{power}
\SetKwFunction{Balance}{balance}
\SetKwFunction{Sinkhorn}{Sinkhorn}
\SetKwFunction{Predict}{predict}
\SetKwFunction{Clean}{clean}
\SetKwFunction{Augment}{augment}
\SetKwComment{Comment}{$\triangleright$ }{}

\SetKwInOut{Input}{input}
\SetKwInOut{Output}{output}
\SetKwFunction{Graph}{graph}
\SetKwFunction{Label}{label}
\SetKwFunction{Softmax}{softmax}
\SetKwFunction{Ce}{ce}

\SetKwFunction{LP}{lp}
\SetKwFunction{Update}{update}
\SetKwFunction{Power}{power}
\SetKwFunction{Balance}{balance}
\SetKwFunction{Sinkhorn}{Sinkhorn}
\SetKwFunction{Predict}{predict}
\SetKwFunction{Clean}{clean}
\SetKwFunction{Augment}{augment}

\Input{ Pre-trained classification network $F$ \\ \& trainable classification head $G$}
\Input{ Training dataset, $D_{\text{train}}$ which contains: \\
Training image for optional training, $\mathbf{x}_{\text{train}}$ \\
Ground truth class for optional training, $y_{\text{train}}$}
\Input{ Input image, $\textbf{x}$}
\Output{ Predicted class, $y$}
\Output{ Predicted segmentation mask $\mathbf{y}$ for $\mathbf{x}$}
\BlankLine

Optionally finetune $F$ and train $G$ with $\mathbf{x}_{\text{train}}$ and $y_{\text{train}}$\\
Predict the class $y$ of input image $\mathbf{x}$ by using $F$\\
Obtain relevance map by applying XAI to result $y$\\
Process the relevance map to obtain a more detailed mask\\
Refine mask with DenseCRF to acquire final segmentation mask $\mathbf{y}$\\

\caption{General Pipeline for ExplainSeg}
\label{alg:adaptive_lp}
\end{algorithm}

\section{Experiments}

\subsection{Setup}
\head{Datasets}

In this research, we conducted experiments on three publicly available medical imaging datasets, each representing a different domain: breast cancer screening, histopathology, and gastrointestinal endoscopy. These datasets include the CBIS-DDSM (Curated Breast Imaging Subset of the Digital Database for Screening Mammography)~\cite{DDSM}, NuInsSeg (Nuclei Instance Segmentation dataset)~\cite{nuinsseg}, and Kvasir-SEG (Polyp Segmentation dataset)~\cite{kvasirseg}.

The CBIS-DDSM dataset is an updated and standardized subset of the DDSM, designed for computer-aided detection and diagnosis in mammography. It contains over 3,000 annotated mammography images across two primary classes: benign and malignant, along with normal cases in some settings. Each case is provided with a corresponding pathology-confirmed label. For our classification training, we used 1,276 images for training and 234 for validation.

The NuInsSeg dataset is a fully annotated benchmark for instance-level nuclei segmentation in H\&E-stained histopathology slides. It consists of 665 high-resolution image tiles with annotations for over 30,000 individual nuclei instances across a variety of tissue types. For our classification-based training, we used 600 images for training and 65 for validation. Although the dataset supports multi-class segmentation, we focused on binary instance segmentation for each nuclei class. All images were normalized using the mean and standard deviation computed from the training set to ensure consistency across staining variations.

The Kvasir-SEG dataset is a polyp segmentation dataset sourced from real-world colonoscopy examinations. It consists of 1,000 colonoscopy images annotated with corresponding binary masks indicating polyp regions. As the dataset contains only positive samples, we split each image into smaller patches and labeled them as positive or negative based on the presence of polyp pixels in the corresponding mask. This approach enabled us to generate a balanced dataset for classification training. We created a training set of 1,852 image patches and a validation set of 254 patches. We computed the mean and standard deviation values from the training images and used these statistics to standardize inputs for both training and inference.

For all datasets, we applied dataset-specific normalization using their respective channel-wise mean and standard deviation values, computed solely from the training subsets to avoid data leakage. Additionally, data augmentation strategies such as horizontal flipping, color jittering, and rotation were applied to improve model robustness and generalization.

Due to the unique structure of the ExplainSeg framework, no information from the ground truth segmentation masks is used during the classification network finetuning stage. Consequently, for the final segmentation mask generation step, both the training and validation images are passed through the trained model to produce predictions, which are then evaluated against the ground truth masks to assess segmentation performance.

\head{Method Variations}

The segmentation pipeline proposed in this study consists of two main stages: the \textbf{explanation stage} and the \textbf{post-processing stage}. For each stage, we introduce two distinct strategies, resulting in four unique combinations: Fusion \& Morphology(Variant A), Fusion \& NCut(Variant B), XAI \& Morphology(Variant C), and XAI \& NCut(XNCut).

In the \textbf{explanation stage}, we explore two alternative approaches to generate meaningful interpretability maps:
    \paragraph{\textbf{XAI-only:}} Employs explainable AI techniques (Integrated Gradients with Noise Tunnel) on the classification network to directly produce heatmaps that indicate regions contributing most to the classification outcome.
    \paragraph{\textbf{Fusion-based:}} Combines the XAI-generated heatmap with intermediate feature maps from the Vision Transformer (ViT) backbone, aiming to enhance spatial precision while retaining semantic relevance.

In the \textbf{post-processing stage}, the relevance maps are transformed into binary segmentation masks using one of the following two techniques:
    \paragraph{\textbf{NCut (Normalized Cut):}} Treats the relevance map as a weighted graph and applies spectral clustering to segment semantically coherent regions.
    \paragraph{\textbf{Morphological Operations:}} Applies classical image processing operations, such as thresholding, dilation, and erosion, to extract connected, object-like structures.

\begin{table*}
\small
\centering
\begin{tabular}{lcccccccccc}
\toprule
\Th{Method Variant}       & \mc{4}{\Th{Components}}                      & \mc{2}{\Th{NuInsSeg}}      & \mc{2}{\Th{CBIS-DDSM}}    & \mc{2}\Th{Kvasir-SEG}       \\
                          & XAI & Fusion & Morph & NCut                        & mIoU\% & Dice\%             & mIoU\% & Dice\%             & mIoU\% & Dice\%       \\
\midrule
Variant A      &      & \ch    & \ch &                          & \tb{13.4} & \tb{22.4}       & 19.0   & 32.3               & 31.1 & 45.6           \\
Variant B            &      & \ch    &   & \ch                           & 12.5   & 20.1             & 11.5   & 25.6               & \tb{32.5} & \tb{45.9} \\
Variant C         & \ch  &        & \ch    &                      & 9.2    & 16.0             & 25.7   & 34.3               & 30.2 & 44.6           \\
XNCut              & \ch  &        &  & \ch                             & 13.1   & 20.6             & \tb{31.2} & \tb{43.7}         & 28.6 & 41.4           \\
\bottomrule
\end{tabular}
\vspace{6pt}
\caption{Ablation study of ExplainSeg variants across three datasets (mIoU\% / Dice\%). Bold indicates best performance per dataset and metric.}
\label{tab:ablation-explainseg}
\end{table*}

Each combination is designed to investigate how different forms of interpretability and refinement influence segmentation performance. We evaluate the effectiveness of these methods across three different datasets, as shown in \autoref{tab:ablation-explainseg}.

\head{Implementation details and hyperparameters} 

Our implementation is based on Python using the PyTorch framework \cite{pytorch}. We benefit from using the Distillation with No Label (DINO) \cite{DINO} Vision Transformer model since it operates in a self-supervised manner making it ideal when labelling is expensive or impractical. 

The 'deitsmall16' pre-trained weights were used to fine-tune the Vision Transformer. Stochastic Gradient Descent (SGD) \cite{sgd} as the optimizer with a learning rate of 0.005, combined with momentum for stability. Additionally, learning rate decay was applied, reducing the learning rate by a factor of 0.1 every 50 epochs to ensure smoother convergence during fine-tuning. To apply XAI to the classification results, both Integrated Gradients and noise tunnelling (with 5 samples) \cite{captum} were implemented to mitigate the impact of noise in both the IG results and the original image.

\subsection{Experimental results}
\head{Baselines Selection}

The core objective of our project is to generate segmentation masks without relying on ground truth annotations. This design choice significantly limits the pool of directly comparable baseline models, as most established segmentation methods require extensive pixel-wise labeled datasets—an approach we intentionally avoid.

While some prior work has explored explainable AI (XAI) methods for segmentation, these are primarily developed for natural image domains and often lack publicly available implementations. As such, they are neither reproducible nor directly applicable to the medical imaging setting of our study, and thus were excluded from experimental comparison.

To assess the performance of our method, we selected two representative baselines: \textbf{TokenCut}\cite{tokencut} and \textbf{MICRA-Net}\cite{micra}. We also incorporated \textbf{MaskCut}, an unsupervised segmentation model hosted within the CutLER repository, which forms the core inference engine in our own pipeline.

TokenCut is an unsupervised segmentation technique that leverages Vision Transformer (ViT) feature tokens and applies spectral clustering to segment images without requiring annotations. Though not specifically designed for medical data, its reliance on high-level semantic features aligns well with our aim of generating segmentation masks driven by meaningful visual cues. TokenCut serves as a strong benchmark for evaluating performance under minimal supervision.

MICRA-Net, by contrast, is a model explicitly designed for medical image analysis. It was pretrained on a domain-specific dataset consisting of F-actin fluorescence microscopy images of hippocampal neurons, as described in the publication: \emph{``Neuronal activity remodels the F-actin based submembrane lattice in dendrites but not axons of hippocampal neurons.''}\cite{micra-data} Although MICRA-Net employs supervised learning, we include it in our comparison to evaluate how a specialized, pretrained model performs in annotation-scarce segmentation tasks.

MaskCut, incorporated as part of the CutLER\cite{cutler} codebase, is a recent unsupervised segmentation method that combines self-supervised ViT features with bipartition-based object discovery. Unlike CutLER’s full pipeline—which includes a clustering and mask refinement stage—MaskCut performs direct bipartition sampling to generate multiple object proposals per image. Despite being developed for general-purpose segmentation, MaskCut proves highly relevant to our study, as it provides multiple candidate masks without requiring pixel-level labels. Its use of CRF-based post-processing and ViT-based semantic representations makes it especially suitable for extracting coarse object boundaries in medical images.

By benchmarking our XAI-driven segmentation approach against TokenCut, MICRA-Net, and MaskCut, we highlight its ability to deliver informative segmentation outputs under limited supervision. This comparative evaluation underscores the novelty and practical relevance of our method in domains—like medical imaging—where annotated data is scarce or expensive to obtain.

\head{Ablation Study}

Table~\ref{tab:ablation-explainseg} compares the four ExplainSeg variants—Fusion \& Morphology(Variant A), Fusion \& NCut(Variant B), XAI \& Morphology(Variant C), and XAI \& NCut(XNCut)—across NuInsSeg, CBIS-DDSM, and Kvasir-SEG using mIoU and Dice metrics.

\textbf{XNCut} achieves the highest mIoU (31.2\%) and Dice (43.7\%) on CBIS-DDSM, the dataset with the highest clinical relevance. It also performs competitively on NuInsSeg and Kvasir-SEG, demonstrating a strong balance between explainability and spatial coherence. This indicates that the integration of explainability-driven pixel relevance with NCut segmentation effectively captures meaningful regions in complex medical images.

Fusion \& Morphology shows solid performance on NuInsSeg and Kvasir-SEG but falls short on CBIS-DDSM, reflecting its limitations in handling highly domain-specific imagery. Fusion \& NCut excels on Kvasir-SEG, achieving the highest Dice (45.9\%) and mIoU (32.5\%), suggesting that NCut’s spatial grouping benefits datasets with clearer anatomical structures and consistent textures. Meanwhile, XAI \& Morphology performs reasonably but does not match the consistency or peak performance of XAI \& NCut(XNCut).

\head{Baseline Comparison}

Table~\ref{tab:method_comparison} presents a comparison between ExplainSeg (XAI \& NCut) and other segmentation methods—TokenCut, MICRA-Net, and MaskCut—across the three datasets.

\begin{table}[ht]
\small
\centering
\setlength\tabcolsep{6pt}
\begin{tabular}{lccc} 
\toprule
{\Th{Method}}           &  {\Th{NuInsSeg}}  & {\Th{CBIS-DDSM}}  & {\Th{Kvasir-SEG}} \\ 
\midrule
\mc{4}{\Th{mIoU}}\\
\midrule
TokenCut \cite{tokencut}               & 5.8              & 1.9               & 23.8              \\
MICRA-Net \cite{micra}                & 7.8               & 12.7               & 15.2                \\
MaskCut  \cite{cutler}                 & 9.3               & 5.4               & \textbf{41.3}                \\
ExplainSeg (XNCut) & \textbf{13.1}             & \textbf{31.2}             & 28.6              \\
\midrule
\mc{4}{\Th{mean Dice}}\\
\midrule
TokenCut \cite{tokencut}               & 10.2              & 3.1               & 28.4              \\
MICRA-Net  \cite{micra}                & 14.1               & 20.8               & 25.3                \\
MaskCut   \cite{cutler}                & 14.3               & 8.5               & \textbf{48.4}                \\
ExplainSeg (XNCut) & \textbf{20.6}             & \textbf{43.7}             & 41.4              \\
\bottomrule
\end{tabular}
\caption{Comparison of segmentation methods on three datasets using mIoU and mean Dice metrics (\%). Bold indicates best performance per dataset and method.}
\label{tab:method_comparison}
\end{table}

\begin{figure}[h]
    \centering
    \includegraphics[width=1\columnwidth]{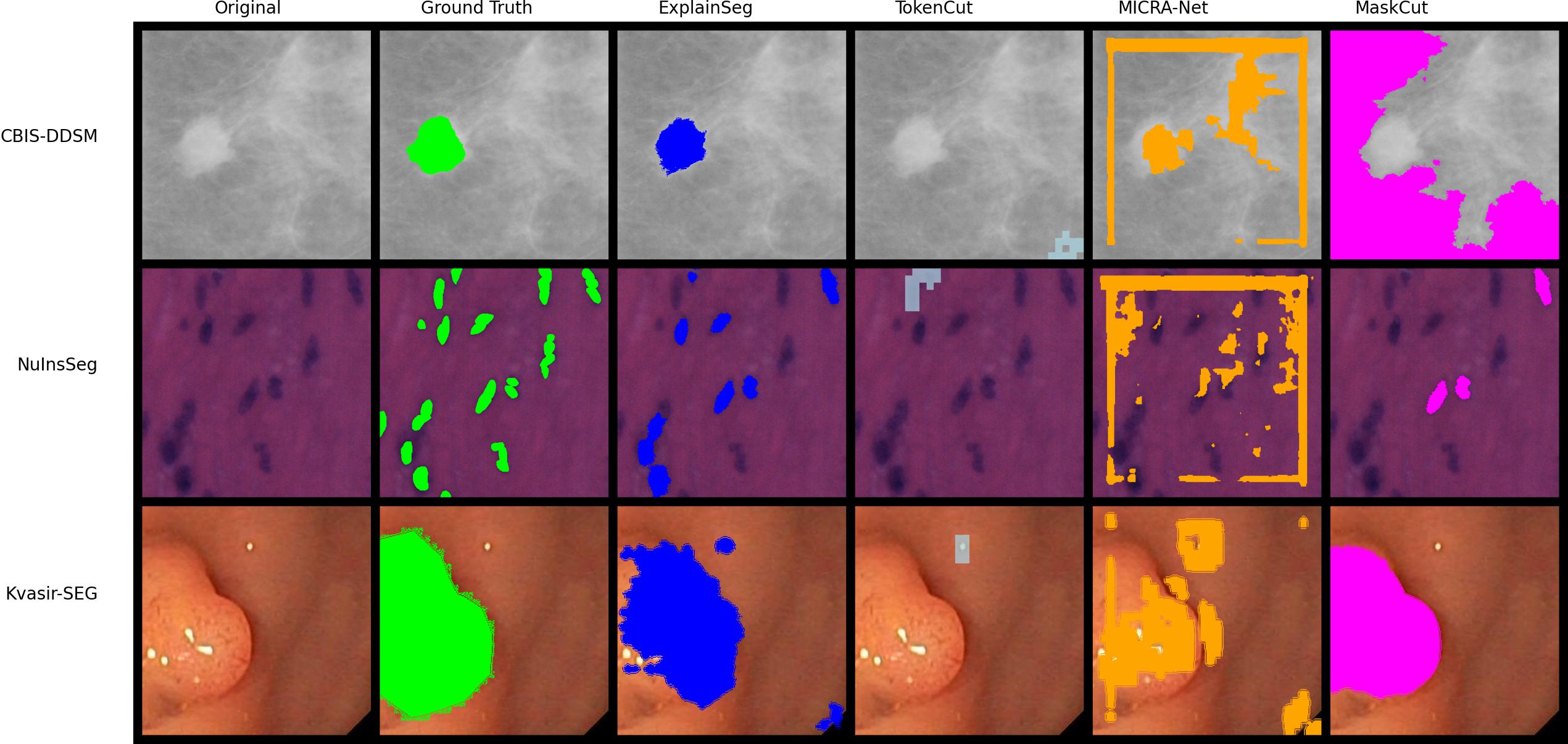}
    \caption{Comparison of results. \\ Top to Bottom: CBIS-DDSM, NuInsSeg, and Kvasir-SEG. \\ Left to Right: Original, Ground Truth, ExplainSeg, TokenCut, MICRA-Net, MaskCut}
    \label{fig:grid_comparison}
\end{figure}

\textbf{ExplainSeg (XAI \& NCut)} consistently achieves the best or near-best segmentation performance across all three datasets. It yields the highest mIoU (31.2\%) and Dice score (43.7\%) on CBIS-DDSM, the mammography dataset that poses substantial challenges due to low contrast and subtle boundaries. On NuInsSeg, which contains microscopy images with densely packed nuclei, ExplainSeg also surpasses TokenCut and MICRA-Net by a clear margin. While it does not attain the top score on Kvasir-SEG, its performance remains highly competitive, underscoring its strong generalizability across diverse imaging modalities.

\textbf{MaskCut}, by contrast, achieves the highest scores on Kvasir-SEG (mIoU 41.3\%, Dice 48.4\%), outperforming all other methods. This is likely attributable to the nature of the dataset, which comprises RGB endoscopic images resembling natural photographs. As MaskCut is pretrained on general-domain image data, its superior performance on Kvasir-SEG is consistent with its inductive bias toward natural image statistics. However, it performs considerably worse on CBIS-DDSM and NuInsSeg, suggesting limited applicability to more specialized or clinical imaging domains.

\textbf{MICRA-Net} performs moderately across all datasets but is consistently outperformed by ExplainSeg, indicating that while it offers a degree of medical-domain awareness, it lacks the fine-grained localization and spatial reasoning facilitated by the integration of XAI and NCut.

\textbf{TokenCut} exhibits the weakest performance overall, particularly on CBIS-DDSM and NuInsSeg, further reinforcing the limitations of general image segmentation techniques when applied to structurally and semantically distinct medical imagery.

These results affirm the effectiveness of ExplainSeg’s XAI-driven relevance maps combined with spectral clustering for producing spatially coherent and clinically meaningful segmentations. Its ability to perform well across varying modalities highlights its robustness and suitability for real-world medical imaging applications.

\section{Conclusions}
\label{sec:conclusion}

This study presents an integrated approach that combines state-of-the-art deep learning techniques with Explainable AI (XAI) to address key challenges in medical image segmentation. Leveraging transfer learning, we fine-tuned a DINO model \cite{DINO} pre-trained with deitsmall16 weights on domain-specific medical imaging data, effectively mitigating the limitations posed by scarce annotated samples. The incorporation of XAI methods enhanced the interpretability of the model, providing critical insights into its decision-making process and fostering greater trust in its predictions.

Our experimental results demonstrate that it is possible to achieve high segmentation accuracy while maintaining clinical interpretability. This balance between performance and transparency is essential for the deployment of AI systems in healthcare, where understanding the rationale behind model outputs is as important as the outputs themselves.

Future research will focus on refining these techniques by investigating more advanced and task-specific XAI methods, expanding the scope to additional medical imaging modalities, and conducting clinical validation. Ultimately, our aim is to develop robust, interpretable AI tools that can assist clinicians in making informed decisions and contribute to improved patient care outcomes.

\bibliographystyle{IEEEbib}
\bibliography{refs}
\end{document}